\documentclass[letterpaper, 10 pt, conference]{ieeeconf}  

\IEEEoverridecommandlockouts                              
  
\usepackage{graphicx}
\usepackage[utf8]{inputenc}

\usepackage{epstopdf}
\usepackage{amsmath}
\usepackage{amssymb}
\usepackage{multirow}
\usepackage{pbox}
\usepackage{algorithm}
\usepackage{algpseudocode}
\usepackage{booktabs}
\usepackage{bm}
\usepackage{url}
\usepackage{esvect}
\usepackage{mathrsfs}
\usepackage{subfigure}

\algtext*{EndIf}
\algtext*{EndFor}

\usepackage{amsthm}
\usepackage{amsfonts}
\usepackage[compact]{titlesec}
\titlespacing{\section}{5pt}{5pt}{5pt}
\titlespacing{\subsection}{5pt}{5pt}{5pt}
\usepackage{cases}
\algnewcommand\algorithmicforeach{\textbf{for each}}
\algdef{S}[FOR]{ForEach}[1]{\algorithmicforeach\ #1\ \algorithmicdo}

\newcommand{\R}{\mathbb{R}}

\newcommand{\E}{\mathbb{E}}

\newcommand\redsout{\bgroup\markoverwith{\textcolor{red}{\rule[0.5ex]{2pt}{0.4pt}}}\ULon}

\usepackage{stfloats}

\usepackage[dvipsnames]{xcolor}

\newtheorem{problem}{Problem}

\theoremstyle{definition}

\DeclareMathOperator*{\argmin}{arg\,min}
\DeclareMathOperator*{\argmax}{arg\,max}

\title{\LARGE \bf Take Your Best Shot: Sampling-Based Planning for\\ Autonomous Photography}

\author{Shijie Gao*$^{1}$, Lauren Bramblett*$^{2}$ and Nicola Bezzo$^{1,2}$
    \thanks{* Co-first authors. Shijie Gao is the corresponding author.} 
    \thanks{Shijie Gao, Lauren Bramblett, and Nicola Bezzo are with the Departments of Electrical \& Computer Engineering$^{1}$ and Systems \& Information Engineering$^{2}$, University of Virginia, Charlottesville, VA 22904, USA. Email: {\tt \{sg9dn, qbr5kx, nb6be\}@virginia.edu}}
}

\renewcommand{\baselinestretch}{0.94}

\begin{document}

\maketitle
\thispagestyle{empty}
\pagestyle{empty}

\begin{abstract}
Autonomous mobile robots (AMRs) equipped with high-quality cameras are revolutionizing the field of autonomous photography by delivering efficient and cost-effective methods for capturing dynamic visual content. As AMRs are deployed in increasingly diverse environments, the challenge of consistently producing high-quality photographic content remains. Traditional approaches often involve AMRs following a predetermined path while capturing data-intensive imagery, which can be suboptimal, especially in environments with limited connectivity or physical obstructions. These drawbacks necessitate intelligent decision-making to pinpoint optimal vantage points for image capture. Inspired by Next Best View studies, we propose a novel autonomous photography framework that enhances image quality and minimizes the number of photos needed. This framework incorporates a proposed evaluation metric that leverages ray-tracing and Gaussian process interpolation, enabling the assessment of potential visual information from the target in partially known environments. A derivative-free optimization (DFO) method is then proposed to sample candidate views and identify the optimal viewpoint. The effectiveness of our approach is demonstrated by comparing it with existing methods and further validated through simulations and experiments with various vehicles.
 
\vspace{3pt}
\noindent\emph{Note---}Code and videos of the simulations and experiments are provided in the supplementary material and can be accessed at {\url{ https://www.bezzorobotics.com/sg-lb-icra25}}.

\end{abstract}


\section{Introduction}
\begin{figure}[t!]
  \centering
  \includegraphics[width=0.9\columnwidth]{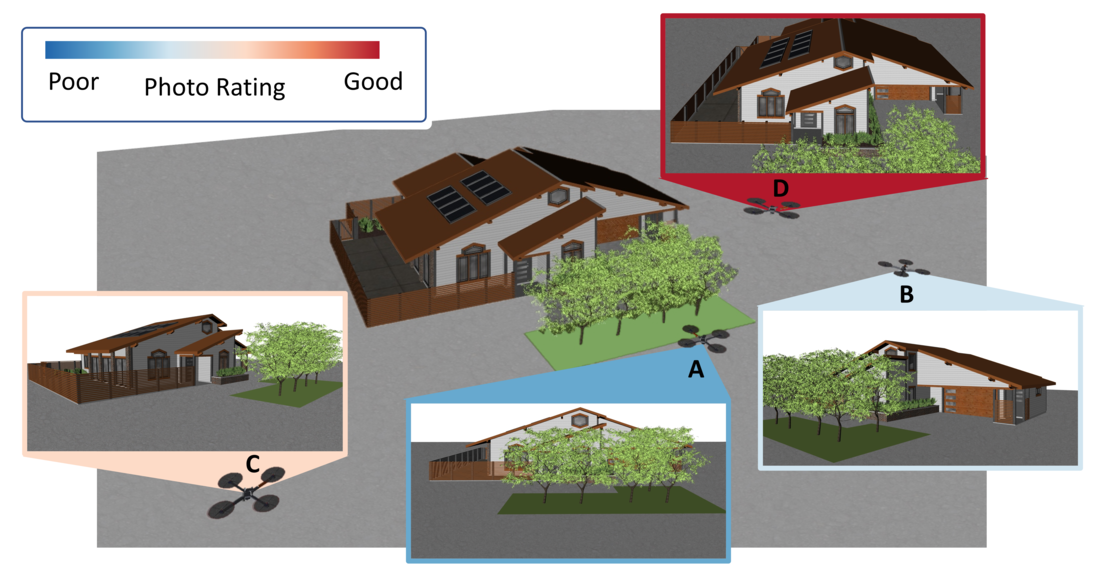}
  \vspace{-8pt}
  \caption{A UAV captures a photograph of a house. Angles A, B, and C represent obstructed or limited views, while a bird's-eye view (D) offers the most comprehensive information with minimal obstruction and distortion.}
  \label{fig:Intro_figure}
  \vspace{-17pt}
\end{figure}

Autonomous mobile robots (AMRs) are becoming increasingly prevalent in industries like manufacturing, healthcare, logistics, real-estate, and entertainment. 
One of the most notable applications in which AMRs play a key role is robotic photography~\cite{zabarauskas2014luke, newbury2020learning} due to their ability to reach view points that would otherwise be inaccessible or even hazardous for humans, e.g., disaster sites, high-rise buildings, or underwater structures~\cite{Joshi2022exploration} or simply to perform tasks more quickly or strategically~\cite{adamson2020designing}.
For example, Fig.~\ref{fig:Intro_figure} depicts a UAV tasked to take the most comprehensive picture of a house. As can be noted by the different viewpoint pictures, some angles are better than others due to the presence of obstacles that occlude the target object of interest. 
This ability to assess and select the best viewpoint (i.e., D in the figure) for capturing critical information is essential to the success of robotic photography tasks, especially in dynamic and cluttered environments.

Despite these advancements, one of the key challenges in automating robotic photography is determining what constitutes a ``good" picture. The evaluation of photo quality is inherently subjective, and even human observers often disagree on what makes an image satisfactory~\cite{byers2003autonomous, montero2015past}. This subjectivity complicates the creation of a formal metric for photo quality assessment. Additionally, current robotic systems often collect large amounts of redundant or irrelevant data, requiring extensive human post-processing to sort and evaluate the captured images.

In this work, we identify two significant gaps in the existing literature: i) the absence of a clear, formal definition of a high-quality picture in the context of robotic photography, and ii) the lack of efficient data collection frameworks that reduce unnecessary images and minimize human involvement in post-processing. Addressing these gaps is crucial to making autonomous photography more practical and scalable in real-world operations. By developing better tools for assessing image quality and optimizing data collection, robots will be able to complete tasks more autonomously, reducing the need for human intervention and increasing overall efficiency.

In our autonomous photography framework, we propose an evaluation metric based on perspective distortion, the scale of a target within the viewing frame, and the estimated target coverage which will allow us to find the best viewpoint. The metric is used in conjunction with a derivative-free optimization (DFO) method which samples the environment to find the best viewpoint. The main contribution of this work is the development of an autonomous photography framework that utilizes computationally efficient Gaussian process interpolation and derivative-free optimization (DFO) to optimize our proposed evaluation metric over uniformly sampled candidate views. This framework enables the runtime capture of high-quality and aesthetically pleasing images of the target in a partially known environment.

\section{Related Work} \label{sec:relatedWork}


This paper addresses a unique challenge in robotic photography, distinct from autonomous inspection. Although both fields aim to capture insightful images, autonomous inspection focuses on detailed visual documentation for precise target reconstruction or digital replication. In contrast, the photographic approach prioritizes obtaining an overall view of the subject.
Our methodology considers that some parts of the target may be missed due to obstructions or occlusions, situations that are less acceptable in inspection contexts where every detail is crucial. Instead, the approach prioritizes capturing the overall essence of the target, similar to human photographers, aiming to gather as much information as possible with few observations.

Next-Best-View (NBV) algorithms, first introduced by Connolly in \cite{connolly1985determination}, focus on the exploration of objects or environments by selecting the most informative next viewpoint for a robot, optimizing this choice based on the robot's goals and constraints. NBV is especially important in 3D object reconstruction, where determining the optimal next viewpoint is crucial. In more recent studies, Naazare et al. \cite{naazare2022online} proposed a weighted multi-objective optimization approach to select NBVs for a mobile robotic arm, while Han et al. \cite{han2022double} used a double-branch network architecture to rank NBVs. Dhami et al. \cite{dhami2023map} extended these ideas by employing two robotic arms for a more efficient reconstruction. However, despite the success of these methods in acquiring dense data, they tend to prioritize collecting information near the target without fully addressing data redundancy or the appearance of the target in the captured photograph.

Another challenge of directly applying NBV-based solutions to autonomous photography is computing the optimal viewpoint in partially known or unknown environments, where unexpected objects can not only obscure the target but also pose safety threats to vehicles. Authors in~\cite{zeng2022deep} navigate robots to optimal viewpoints in known environments, while the authors in \cite{bircher2016receding} investigate a receding horizon NBV, utilizing a random tree method to guide the robot along a path in an unknown environment. \cite{almadhoun2019guided} proposes a guided NBV approach for large-scale 3D reconstruction, which requires a rough global scan prior to a detailed NBV inspection. \cite{monica2019humanoid} computes NBV based on the map built during frontier exploration. 
In contrast, our work focuses on applications where the target location is known but the surrounding environment is unknown such as real-estate or surveillance. Given the target location, the robot dynamically updates the optimal viewpoint to efficiently adapt to changes in the environment.

\section{Problem Formulation} \label{sec:problemFormulation}

\vspace{-2pt}

For typical photography missions, target information is known, but the surroundings can be uncertain. We assume the robot knows the target's location and dimensions $\mathbf{T}$. Obstacles, represented as an occupancy map $\mathcal{M}$, can be known or unknown $\mathcal{O}$. The vehicle should update its viewpoint when detecting obstacles that block the target. Equipped with range sensors (LiDAR, sonar, cameras), the robot can recognize obstacles and measure its distance to the target.
The goal is to maximize the information captured in the photo while reducing the number of
pictures needed during operations. To achieve this, we formally define the problems as follows:
 
\vspace{-3pt}
\begin{problem}\label{problem1}{\bf{\emph{Metrics of the Best View for Capturing Target:}}}
Consider the location and dimensions of the target represented by the coordinate set $\mathbf{T}$ and a set of initially unknown obstacles that are discovered at runtime and represented by an occupancy map $\mathcal{M}$. The goal is to find an evaluation metric $G(\mathbf{T}, \mathcal{M}, \mathbf{P})$ that scores the sampled viewpoints $\mathbf{P}$ for the vehicle to visit at runtime.
\end{problem}

\begin{problem}\label{problem2}{\bf{\emph{Max-Info Path Planning:}}} 
Given a partially-knownn environment and metrics for picture evaluation, generate a path that minimizes the time to the viewpoint while ensuring the robot's safety. The path planner should also accommodate changes in the environment, as new obstacles may appear as the vehicle approaches the best viewpoint.
\end{problem}
\section{Methodology} \label{sec:methodology}
\begin{figure}[ht!]
  \centering
  \includegraphics[width=\columnwidth]{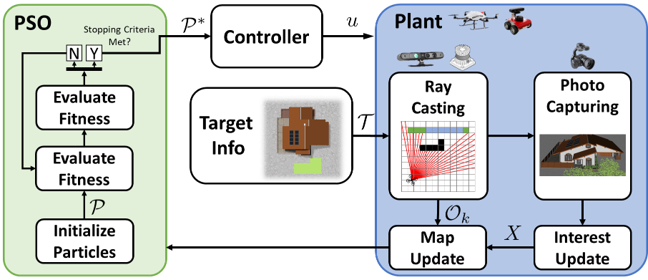}
  \vspace{-20pt}
  \caption{Diagram of the proposed approach.}
  \label{fig:methodology_block_diagram}
  \vspace{-15pt}
\end{figure}

The framework aims to autonomously capture the entire target from as few viewpoints as possible. The camera position ensures the target occupies a specific proportion of the frame, minimizing distortion and obstructions. If the target is not fully visible in one frame, the robot must decide autonomously to focus on one part of the target and subsequently move to capture the remainder of the target. In this work, we propose an objective function that formally defines the quality of a photograph and a method to capture quality photographs of the entire target with as few pictures as possible. 
Fig.~\ref{fig:methodology_block_diagram} describes a method for guiding an autonomous mobile robot during a mission in a partially-known environment. Starting from a known point, the robot utilizes a depth sensor to update its map and interest in unseen portions of the target while generating candidate views. These views are scored for quality, and particle swarm optimization (PSO) is employed to determine the optimal camera position, though any DFO method could be applied. The robot continuously re-evaluates its viewpoint as it moves, updating its target coverage using Gaussian process interpolation once the optimal point is reached and a photo is taken. Further details on metrics and view evaluation follow.


\subsection{Photo Evaluation Metric}
\begin{figure}[ht!]
  \centering
  \includegraphics[width=0.78\columnwidth]{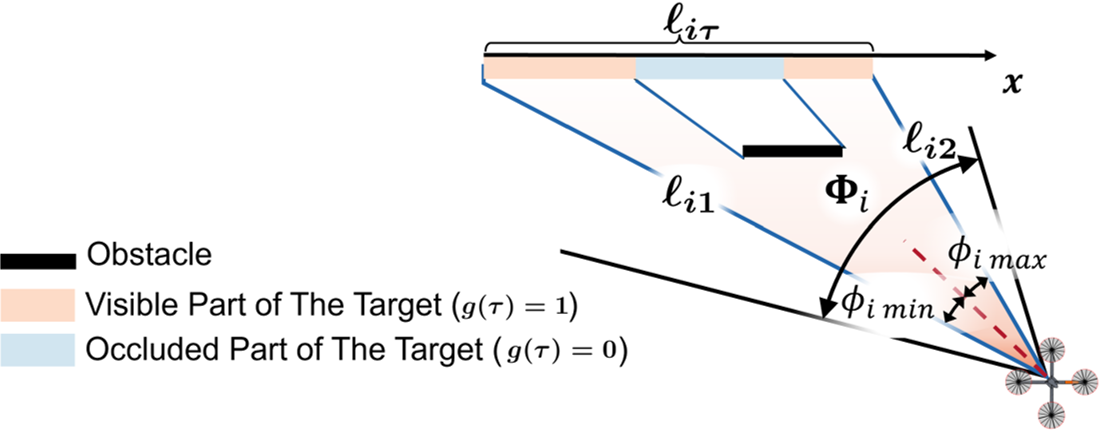}
  \vspace{-8pt}
  \caption{Pictorial depiction of the evaluation metric for a 2D space.}
  \label{fig:methodology_2d_example}
\end{figure}

\begin{figure}[ht!]
  \centering
  \includegraphics[width=0.8\columnwidth, height=3.5cm]{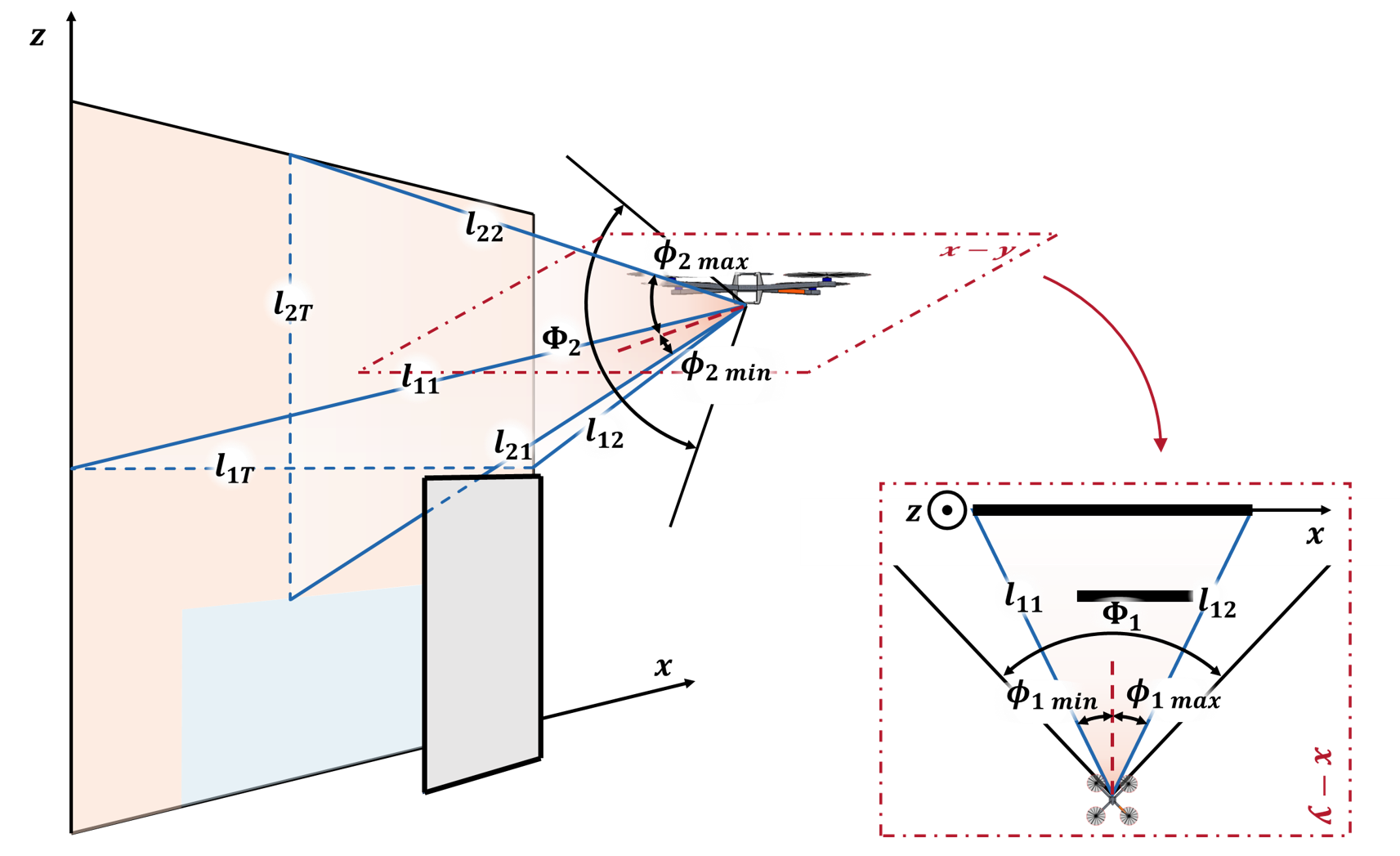}
  \vspace{-10pt}
  \caption{Pictorial depiction of evaluation metric for a 3D space.}
  \label{fig:methodology_3d_example}
  \vspace{-10pt}
\end{figure}

First, we introduce the metric to determine the ideal position to capture an image of a desired target which were chosen based on careful discussion with our research sponsors in the real-estate industry. To aid the reader's comprehension, we begin with a visual illustration of the metric using a two-dimensional example in Fig.~\ref{fig:methodology_2d_example} followed by an example in Fig. \ref{fig:methodology_3d_example} that showcases the expansion of the 2D metrics to 3D. We consider a discretized target $\mathbf{T}\in\R^{m\times n_d}$ where $n_d$ is the dimension of the target and defined by $m$ coordinates. 
In Fig.~\ref{fig:methodology_2d_example}, a UAV is positioned slightly behind an obstacle, obscuring its view of the full target. 
For the visible portion of the target $\mathbf{T}_v\subseteq \mathbf{T}$ and for $\tau\in\mathbf{T}_v$, $g(\tau,\bm{p})=1$ indicates that the visible portion has not been captured previously and is visible from viewpoint $\bm{p}$, while $g(\tau,\bm{p})=0$ indicates that the region is obscured or has already been observed in the past, making the interest in that section low or unimportant to recapture. 
We define the optimal viewpoint  $\bm{p}^*$ as follows: 
\begin{equation}
   \bm{p}^* = \underset{\bm{p}}{\argmax} \ \ G(\mathbf{T}, \mathcal{M}, \bm{p}) \label{eq:optFunc}
\end{equation}
\vspace{-5pt}
\begin{equation}
    \textrm{where}\;\; G(\mathbf{T}, \mathcal{M}, \bm{p}) = \gamma_d \cdot \gamma_s \cdot \sum_{\tau\in\mathbf{T}} g(\tau,\bm{p}).\nonumber
\end{equation}
\noindent The function $g(\tau,\bm{p})$ signifies the level of interest that remains for any portion of the target after viewing from viewpoint $\bm{p}$. As shown in Fig.~\ref{fig:methodology_2d_example}, the visible portion of the target is a function of the position $\bm{p}$ and the occupancy map $\mathcal{M}$. The variables $\gamma_d$ and $\gamma_s$ are discount factors associated with utility functions $U_d$ and $U_s$ to assess the quality of a viewpoint given its respective camera parameters, position, and environmental obstructions. These factors are essential in determining the robot's choice of optimal viewpoints, as they influence the amount of distortion that can be accepted in the pictures and the amount of the target that should occupy the image. The evaluation of these factors will be discussed in more detail in the following sections. 

\vspace{-8pt}
\vspace{-15pt}

\subsubsection{Perspective Distortion Discount Factor} 
Directly facing the target surface is often the ideal view to maximize the amount of information in one photograph and create a consistent view of the target~\cite{friedman2006automated}. In fields such as product photography, architecture, or real estate where precise measurements and representations are required, 
distorted structure is less preferred. We introduce a {\em perspective distortion} discount factor that penalizes views that cause a distortion of the target. 
The target distortion in our approach is formally defined as follows:
\vspace{-5pt}
\begin{equation}
    \gamma_d = \prod_{i=1}^{n_d-1} U_d\left(\frac{|\ell_{i2}-\ell_{i1}|}{\ell_{i\mathbf{T}}},\mathbf{q}_d\right)
    \label{eql:perspective_2d}
\end{equation} 
where $\ell_{i1}$ and $\ell_{i2}$ represent the maximum distances from the optical center associated with the intersection of the camera frame and the target. The variable $\ell_{i\mathbf{T}}$ is the length of the target captured along dimension $i$. $U_d$ represents the user-defined utility function associated with the perspective distortion, penalizing unbalanced target captures, and $\mathbf{q}_d$ is a vector of parameters for the user-defined function. For example, in a 2D scenario if $\ell_{11}=\ell_{12}$, the view of the vehicle's camera is perpendicular to the target, minimizing perspective distortion and maximizing $\gamma_d$. If $\ell_{11}>>\ell_{12}$ or $\ell_{11}<<\ell_{12}$, the vehicle is positioned far to the left or right of the target, respectively, maximizing the perspective distortion and resulting in a small $\gamma_d$.  
By incorporating this factor into the optimization process, the proposed method can effectively balance the trade-off between the ideal view and the available view, improving the photo quality. 

\subsubsection{Scale Discount Factor}
In photography, the term {\em scale} refers to the relative size of the target compared to other objects in the frame. A favorable picture places the target in the foreground, occupying a percentage $\beta$ of the frame. Proper scaling is crucial for accurately representing the target. 
The scale discount factor $\gamma_s$ is used to encourage capturing images where the target occupies the desired percentage of the frame, determined by $\beta$. 
We use the following equation to calculate this factor:
\begin{equation}
    \gamma_s =\prod_{i=1}^{n_d-1}  U_s\left(\frac{(\phi_{i \max}-\phi_{i \min})}{\Phi_i},\beta,\mathbf{q}_s\right)
    \label{eq:gamma_s}
\end{equation}
where $\phi_{i \min}$ and $\phi_{i \max}$ are the minimum and maximum angles of the intersection between the target plane and center of the frame along dimension $i$ and $\Phi_i$ is the field of view of the camera along dimension $i$. We denote $U_s$ as the user-defined utility function associated with the scale of the target in the frame and $\mathbf{q}_s$ as a vector of parameters for the user-defined function. In \eqref{eq:gamma_s}
, if $\phi_{i\min} = -\Phi_i/2$ and $\phi_{i\max}=\Phi_i/2$, the target would occupy $100\%$ of the frame along dimension $i$. The utility function $U_s(\cdot)$ would score this percentage based on how close it is to the desired percentage $\beta$. 
Fig.~\ref{fig:methodology_2d_example} and Fig.~\ref{fig:methodology_3d_example} present a physical representation to illustrate this concept. 




\subsection{Candidate View Evaluation}

The quality of pictures taken at viewpoints is evaluated based on the vehicle's knowledge of the environment. The vehicle updates the environment's occupancy map $\mathcal{M}$ using recursive Bayesian updates~\cite{asgharivaskasi2021active}, which helps re-evaluate viewpoints as it moves, as view quality can change with new observations. To manage the complexity of determining optimal views in cluttered environments, ray-casting is used to assess viewpoint quality and reduce computational load.

\begin{figure}
    \centering
    \includegraphics[width = 0.21\textwidth]{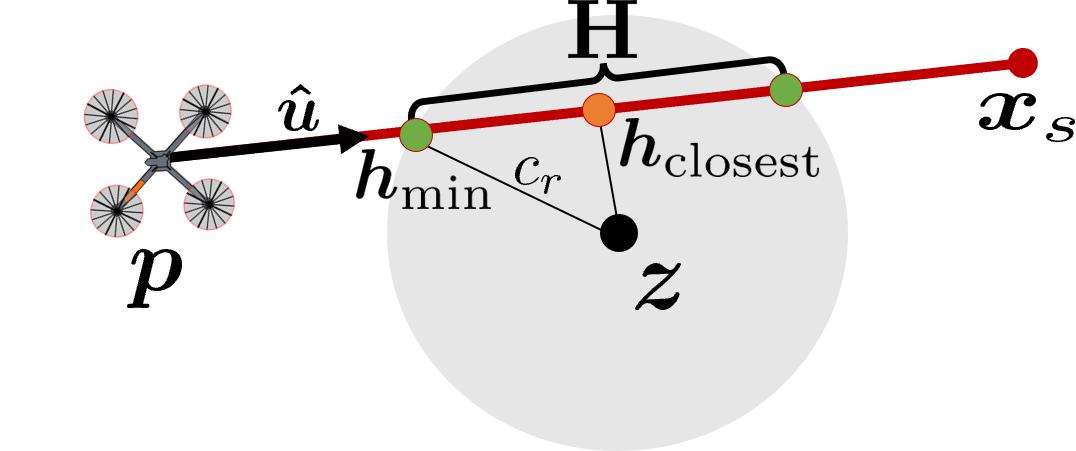}
    \vspace{-5pt}
    \caption{Pictoral depiction of a ray-sphere intersection with coordinate $\bm{z}$. The intersection point $\bm{h}_{\min}$ is used to update the coordinate $\bm{x}_s$.}
    \label{fig:rayCastExample}
    \vspace{-20pt}
\end{figure}

Ray-casting, a classical technique in computer vision, allows for discrete modeling of complex interactions and quality estimations~\cite{vasquez2009view}. Following~\cite{vasquez2013hierarchical}, we use ray-casting and \eqref{eq:optFunc} to estimate the view quality. In our approach, a geometric ray-sphere transformation is employed to efficiently estimate interactions with obstacles and the target. The set of coordinates terminating at the vehicle's maximum observation range $d_\text{max}$ is derived from a set of unit vectors $\mathbf{U}$, defined as:
\begin{equation}
    \mathbf{X}_s = \{\bm{p} + d_\text{max}\bm{\hat{u}} \ \ | \ \bm{\hat{u}}\in\mathbf{U}\}.
\end{equation}
We also define a radius $c_r$ to check each ray for intersections with the discretized target space $\mathbf{T}$ or obstacle coordinates. As shown in Fig.~\ref{fig:rayCastExample}, we assess the intersection of a coordinate of interest $\bm{z}$ (e.g., target or obstacle) with each ray from the viewpoint in the set $\mathbf{U}$. Using a geometric approach, we project the line segment from the viewpoint $\bm{p}$ to the coordinate along the vector $\bm{\hat{u}}$ for the coordinate $\bm{x}_s$. Algorithm~\ref{alg:raytracing} overviews the ray-tracing procedure for a coordinate $\bm{z}$.
\begin{algorithm}
  \caption{\strut Ray-casting algorithm}\label{alg:raytracing}
  \begin{algorithmic}[1]
  \Require{$\bm{x}_s; \ \hat{\bm{u}}; \ \mathbf{T}$}
    \State $y_s = 0$
    \State $d_\text{terminal} = ||\bm{x}_s-\bm{p}||_2 $
    \State $d_z = (\bm{z}-\bm{p})\cdot\bm{\hat{u}}$
    \If{$d_z>0$}
        \State $\bm{h}_\text{closest} = \bm{p} + d_z\bm{\hat{u}}$ \Comment{Closest point along ray to $\bm{z}$}
        \State $d_\text{center} = ||\bm{h}_\text{closest}-\bm{z}||_2$
        \If{$d_\text{center}<c_r$}
            \State $ d_{\text{chord}} = \sqrt{c_r^2-d_{\text{center}}^2} $
            \State $\mathbf{H} = \bm{h}_\text{closest}\pm d_{\text{chord}}$ 
            \State $\bm{h}_{\min} = \underset{\bm{h}_s\in \mathbf{H}}{\argmin} \ ||\bm{h}_s-\bm{p}||_2 $
            \State $d_\text{intersect} = ||\bm{h}_{\min}-\bm{p}||_2$
            \If{$d_\text{intersect}<d_\text{terminal}$}
                \State $d_\text{terminal} = d_\text{intersect}$
                \State $\bm{x}_{s} \leftarrow \bm{h}_{\min}$
                \If{$\bm{z}\in \mathbf{T}$}
                    \State $y_s \leftarrow 1 $
                \EndIf
            \EndIf
        \EndIf
    \EndIf
    \State \text{return} $\bm{x}_s,y_s$
  \end{algorithmic}
\end{algorithm}
Given that this algorithm can find intersections with target coordinates and obstacles, each $\bm{x}_s\in\mathbf{X}_s$ is updated using Algorithm~\ref{alg:raytracing}, and whether each $\bm{x}_s$ is associated with the target is stored, denoted logically by $y_s\in Y_s$.

Subsequently, we can find the values in \eqref{eql:perspective_2d} and \eqref{eq:gamma_s} that intersect the target. $\phi_{i\min}$ and $\phi_{i\max}$ represent the minimum and maximum angles between the normal vector and the vectors that intersect the target along the $i^{\text{th}}$ dimension. Similarly, $\ell_{i1}$ and $\ell_{i2}$ are determined by the distance to the intersection point along dimension $i^{\text{th}}$. This approach allows us to compute the score in \eqref{eq:optFunc} discretely, increasing the computation speed for each candidate view. 
 
\subsection{Target Coverage Estimation using GP Interpolation}\label{sec:approach-GPI}

The exact coverage for any target in 2D or 3D space is a binary indicator for all coordinates of whether a coordinate of the target has been captured or not. However, this process can be expensive for a vehicle with limited computing capability for any sized target, especially considering that we would need an infinite number of rays to exactly compute the coverage. For this reason, we use sparse ray-casting and Gaussian process interpolation, utilizing induction points based on the target location.

Gaussian process interpolation is a probabilistic framework for interpolation and provides a natural way to quantify uncertainty for a static set of target points \cite{yel2020gp,corah2019communication}. Given a finite and sparse number of sampling points from the ray intersections, we use Gaussian process interpolation to estimate the probability of a target coordinate $\bm{x}\in\mathbf{X}$ being observed at non-sampled target points. 

\begin{figure}[t!]
    \centering
    \includegraphics[width=0.9\columnwidth]{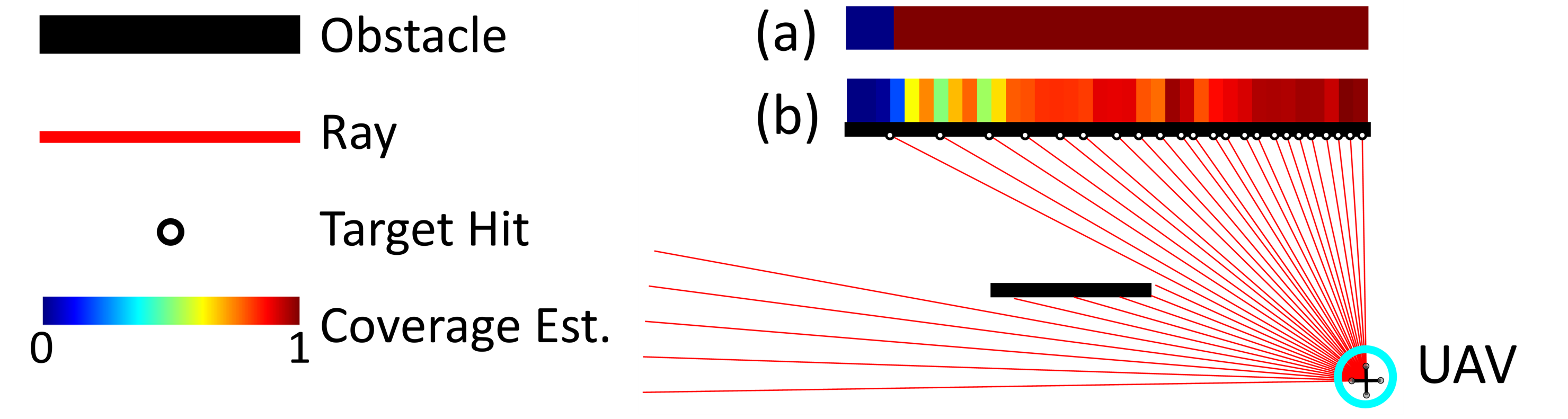}
    \vspace{-10pt}
    \caption{An example of GP interpolation for 2D target visibility. The color bar (a) shows true target coverage and (b) shows GP interpolated coverage.}
    \vspace{-13pt}
    \label{fig:2D_gp_showcase}
\end{figure}

Consider an initial GP, as presented in \cite{williams2006gaussian}
\begin{equation}
    f(\mathbf{X})\sim \mathcal{GP}(\mu(\mathbf{X}),k(\mathbf{X},\mathbf{X}'))
\end{equation}
characterized by a mean and covariance function:
\begin{gather}
    \mu(\mathbf{X}) = \E[f(\mathbf{X})]\\
    k(\mathbf{X},\mathbf{X}') = \E[(f(\mathbf{X})-\mu(\mathbf{X}))(f(\mathbf{X}')-\mu(\mathbf{X}'))]
\end{gather}
where $f(\mathbf{X})$ is functionally interpreted in this application as whether a target coordinate $\bm{x}\in \mathbf{X}$ has been captured by the robot (i.e., a zero if the coordinate has not been observed and a one if the coordinate has been captured before). For interpolation of the target space, the widely used radial basis function (RBF) kernel is employed to capture the spatial relationship between two points and likewise interpolate their value using the Gaussian process regression with the kernel equation formulated as:
\begin{equation}
    k(\mathbf{X},\mathbf{X}') = \sigma_f^2\exp\left(-\frac{||\mathbf{X}-\mathbf{X}'||^2}{2\sigma_l^2}\right).
    \label{eq:kernel}
\end{equation}
where $\sigma_l \in[0,1]$ and $\sigma_f \in[0,1]$ are two hyperparameters. In this work, we tuned the hyperparameters by minimizing the negative log-likelihood of the training data. 
In this application, the variable $\mathbf{X}\in\R^{m\times n_d}$ is a set of $m$ sampled target coordinates from the ray-casting discussed in the previous section. We can estimate the target coverage as the difference between the prior target coverage $\mu(\mathbf{X})$ and the estimated target coverage using the posterior mean, $\mu^*(\mathbf{X})$, from a new viewpoint $\bm{p}$ given the set of sample target points $\mathbf{X}_s\in\R^{s\times n_d}$. As such, the computational complexity of the target coverage is significantly reduced by the new equation
\begin{equation}
    G(\mathbf{X},\bm{p}) = \gamma_d\cdot\gamma_s\cdot\sum_{j=1}^{m}(\mu_j^*(\mathbf{X})-\mu_j(\mathbf{X})).
\end{equation}
where the subscript on $\mu$ denotes the element of the mean value for coordinate $j$ and we compute $\mu^*$ as
\begin{equation}
    \mu^*(\mathbf{X}) = k(\mathbf{X},\mathbf{X}_s)\left[k(\mathbf{X}_s,\mathbf{X}_s)+\sigma_n^2 I\right]^{-1}Y_s.
\end{equation}
The variable $\sigma_n$ is a constant in this case representing sensor noise and $Y_s$ is the binary 1D matrix signifying if the sample terminal ray-cast coordinates in $\mathbf{X}_s$ were in range of a target coordinate. With this belief update, we estimate the probability of additional coverage of the target from a candidate viewpoint $\bm{p}$ and \eqref{eq:optFunc} can be efficiently computed for each candidate viewpoint $\bm{p}$. Fig.~\ref{fig:2D_gp_showcase} showcases the estimated view coverage using ray tracing and GP in 2D. However, computing every candidate view in the environment is intractable at runtime, especially given that \eqref{eq:optFunc} may have a non-linear or discontinuous solution space. To overcome this limitation, we incorporate particle swarm optimization to dynamically determine the optimal viewpoint in real-time.

 \subsection{Candidate View Particle Swarm Optimization}\label{sec:approach-pso}
In our autonomous photography framework, a candidate viewpoint is represented as a point in a $N$-dimensional solution space. A swarm of particles is used to represent potential viewpoints. Each particle, $\bm{p}_i(t)$, is the coordinate of the $i^{\text{th}}$ candidate view at time $t$, and its velocity components are represented as $\bm{v}_i(t)$. The coverage $G$ is calculated for each viewpoint and the particles migrate toward the particle with the highest evaluation score.

 At each time $t$, the velocity of a particle is perturbed by the weighted sum of its personal best view and global best view, resulting in an updated velocity and position. To update the particle, we use the following equations:
 \vspace{-15pt}

 \begin{align}
     \bm{v}_i(t+1) & = w\bm{v}_i(t) + r_1c_1(\bm{b}_{i}(t)-\bm{p}_i(t)) \\ 
     &+r_2c_2(\bm{g}(t)-\bm{p}_i(t))\nonumber\\
    \bm{p}_i(t+1) &= \bm{p}_i(t)+\bm{v}_i(t+1)
\end{align}
where $\bm{b}_{i}(t)$ and $\bm{g}(t)$ 
 are the global best position evaluated by the particles and the global best position of all particles, respectively. The acceleration coefficients $c_1$ and $c_2$ are non-negative constants, which weigh the influence of the personal and global bests in the search process. The inertia weight $w$ balances the local and global exploration of the particles in the search space \cite{zhang2013robot}. To prevent divergence, $w$ must be between 0 and less than 1. Adjusting coefficients $c_1$ and $c_2$ influences particles to explore around local optima (when $c_1>c_2$) or global optima (when $c_2>c_1$). This can be modified during the heuristic process for exploration or exploitation. Variables $r_1$ and $r_2$, random numbers between 0 and 1, perturb particles to encourage exploration.

\begin{figure}[t]
    \subfigure[]{
    \includegraphics[width = 0.28\columnwidth]{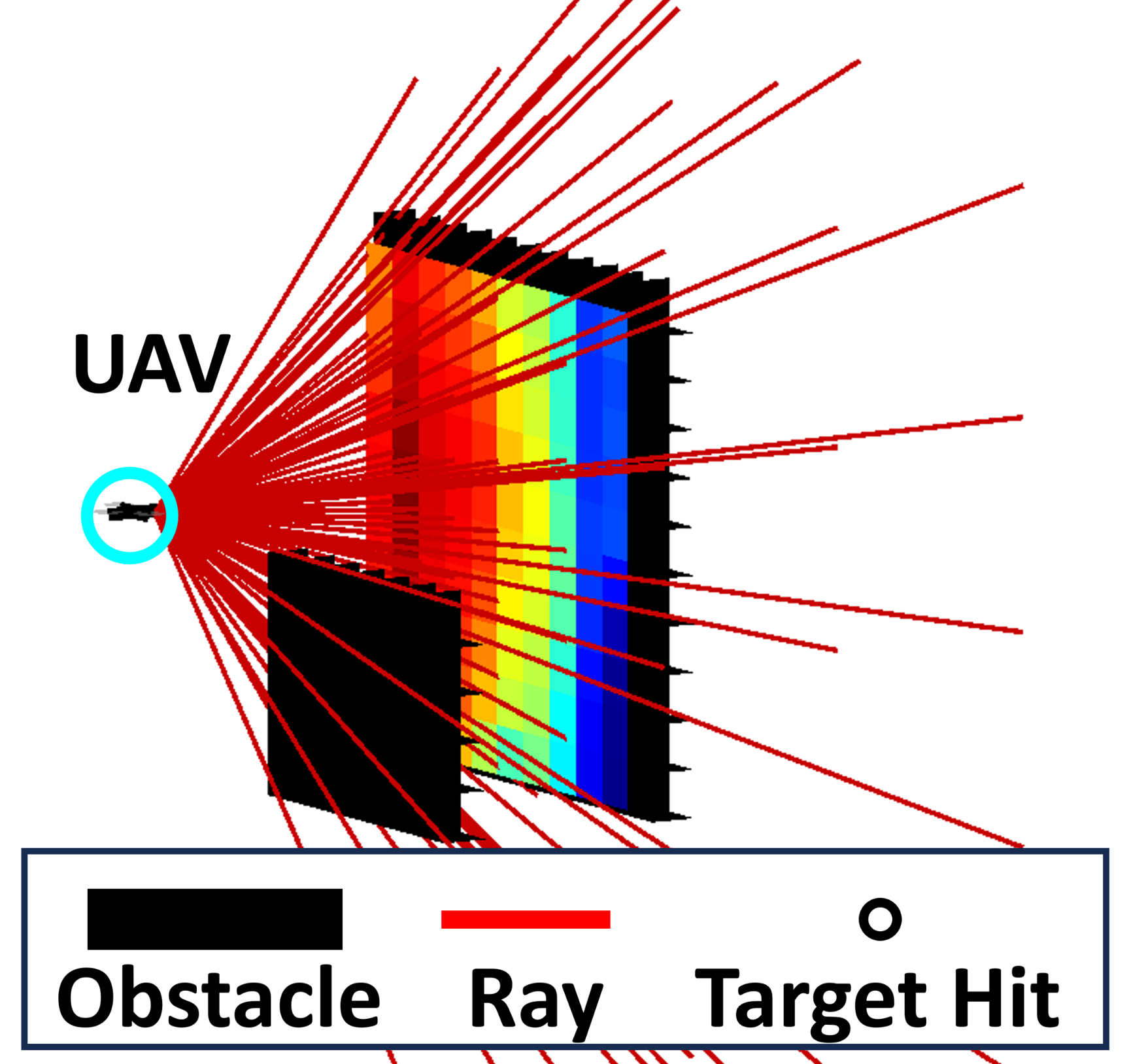}
    \label{fig:3d_gp_showcase}
    }%
    \subfigure[]{
    \includegraphics[width =0.31\columnwidth]{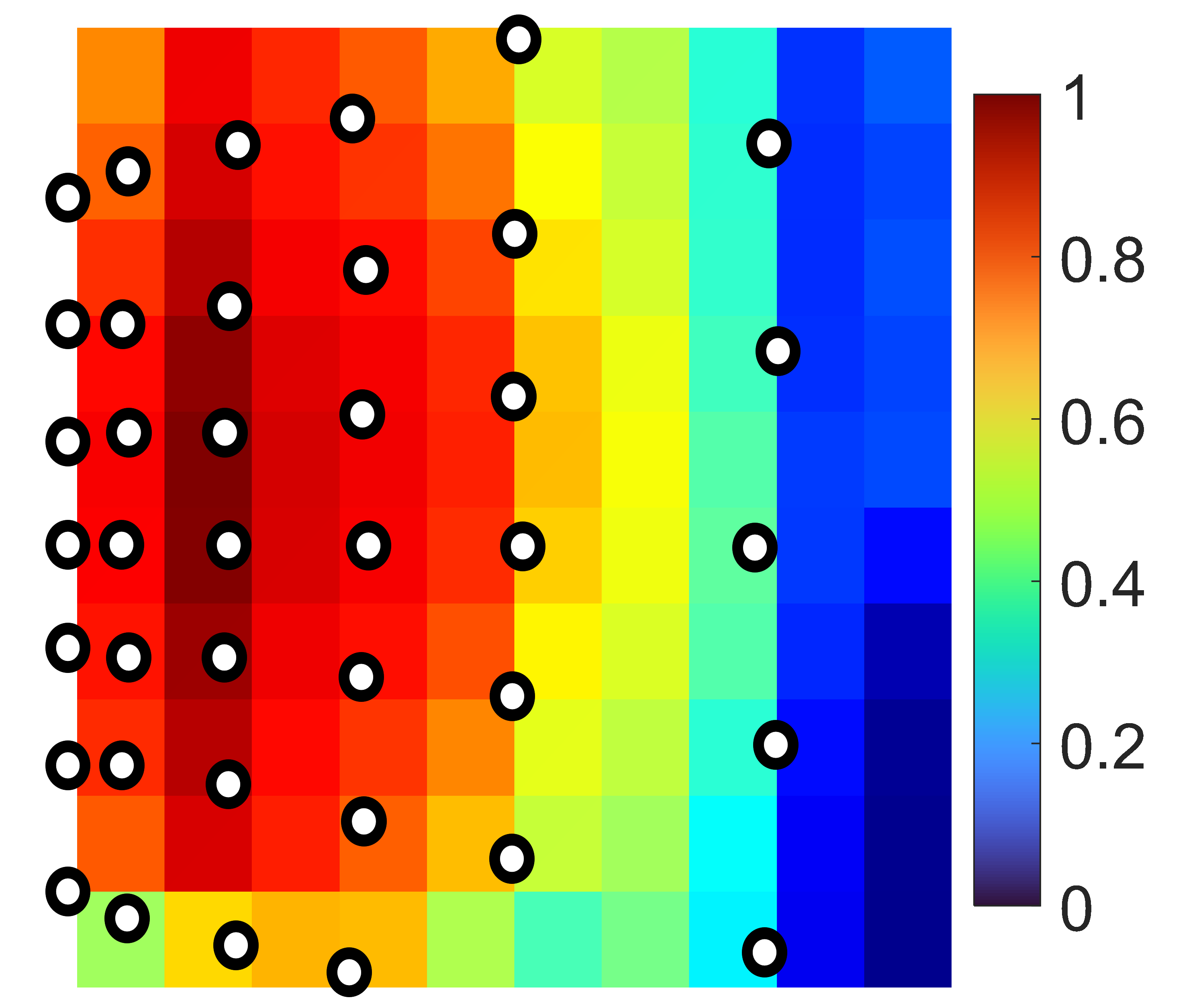}
    \label{fig:3d_gp_estimation}%
    }%
    \subfigure[]{
    \includegraphics[width =0.31\columnwidth]{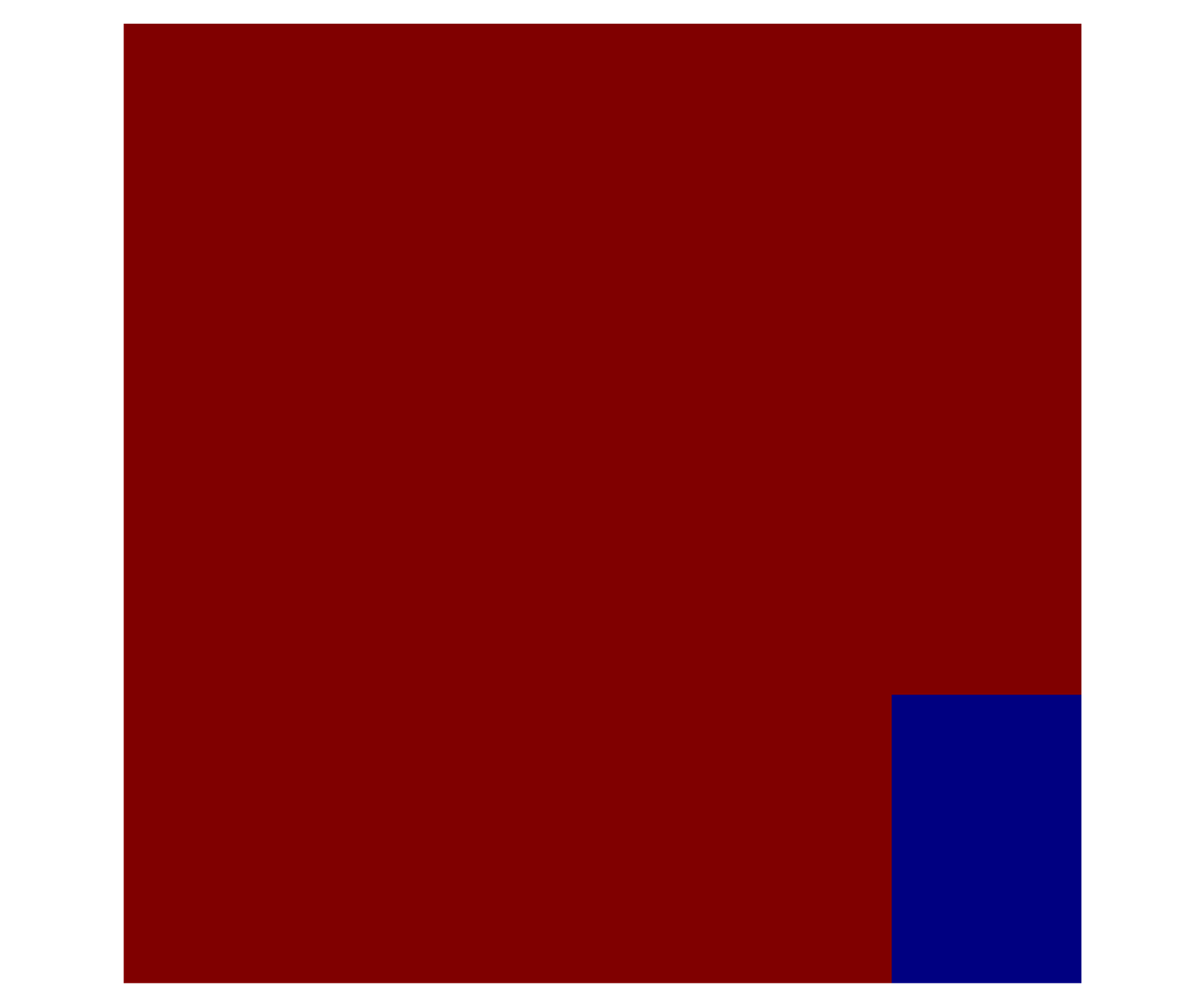}
    \label{fig:3d_gp_groundtruth}\\
    }%
    \vspace{-10pt}
\caption{An example of using GP to estimate the visibility of the target in 3D space. (a) shows the environment, (b) shows the estimated GP result from the ray hits. (c) provides the ground truth visibility.}
\label{fig:3d_gp}
\vspace{-17pt}
\end{figure}

 Fig.~\ref{fig:3d_gp} illustrates our 3D approach where the vehicle detects an obstacle obstructing the target and propagates particles using the updated occupancy map. As the vehicle moves toward the best viewpoint, the particles iteratively search for the optimal view. Upon reaching the best viewpoint, the vehicle captures a photo and updates the observed target areas, as shown in Fig.~\ref{fig:3d_gp_estimation}. 
 The obstacle shifts the optimal viewpoint, causing the right side to be marked as unseen and targeted for capture in future iterations.

\section{Simulations}\label{sec:simulation}
To demonstrate the robustness of the proposed method, we conducted extensive simulations using different vehicles equipped with sensors of varying capabilities in diverse environmental contexts. In these MATLAB simulations, the vehicle operates in a partially-known environment defined by operational boundaries, target position, and dimensions. The vehicle is equipped with two primary sensor types: 1) a depth sensor for obstacle detection (e.g., RGB-D camera, Lidar, etc.), and 2) an optical camera.

In the 2D simulation, the vehicle operates in a 100m$\times$100m grid world with a resolution of 1m and uses hybrid A* for path planning. The objective is to capture the four sides of a building measuring 30m$\times$15m. Various shapes of unknown obstacles are placed in front of three sides of the building. Both the depth sensor and the camera have a field of view (FOV) of $\Phi=[-\frac{\pi}{4}, \frac{\pi}{4}]$, and the range of the RGB-D sensor is $r_{max}=25$m.

We set $\beta=0.8$, and the utility functions are designed in the style of logistic functions:
\begin{equation}
    U(x, \mathbf{q}) = q_1 + \frac{q_2}{1+e^{q_3(x+q_4)}}, \quad \mathbf{q} = [q_1, ...,q_4]
\end{equation}
We chose logistic functions for their rapid decline, allowing effective differentiation between preferred values and less desirable ones. For the utility function $U_d$, which penalizes as $x$ approaches $1$, we select $\mathbf{q}_d = [0.3,0.7,20,-0.75]$. For $U_s$, we construct a piecewise function with $\mathbf{q}_s = [0,1,-20,-0.5]$ for $x\in[0, \beta]$ and $\mathbf{q}_s = [0,1,30,-1]$ for $x\in(\beta, 1]$. These functions ensure the utility’s peak aligns with the desired proportion $\beta$, reducing distortion and preventing view intersection with the frame edges~\cite{zabarauskas2014luke}.

\begin{figure}[h]
    \centering
    \includegraphics[width=0.95\columnwidth]{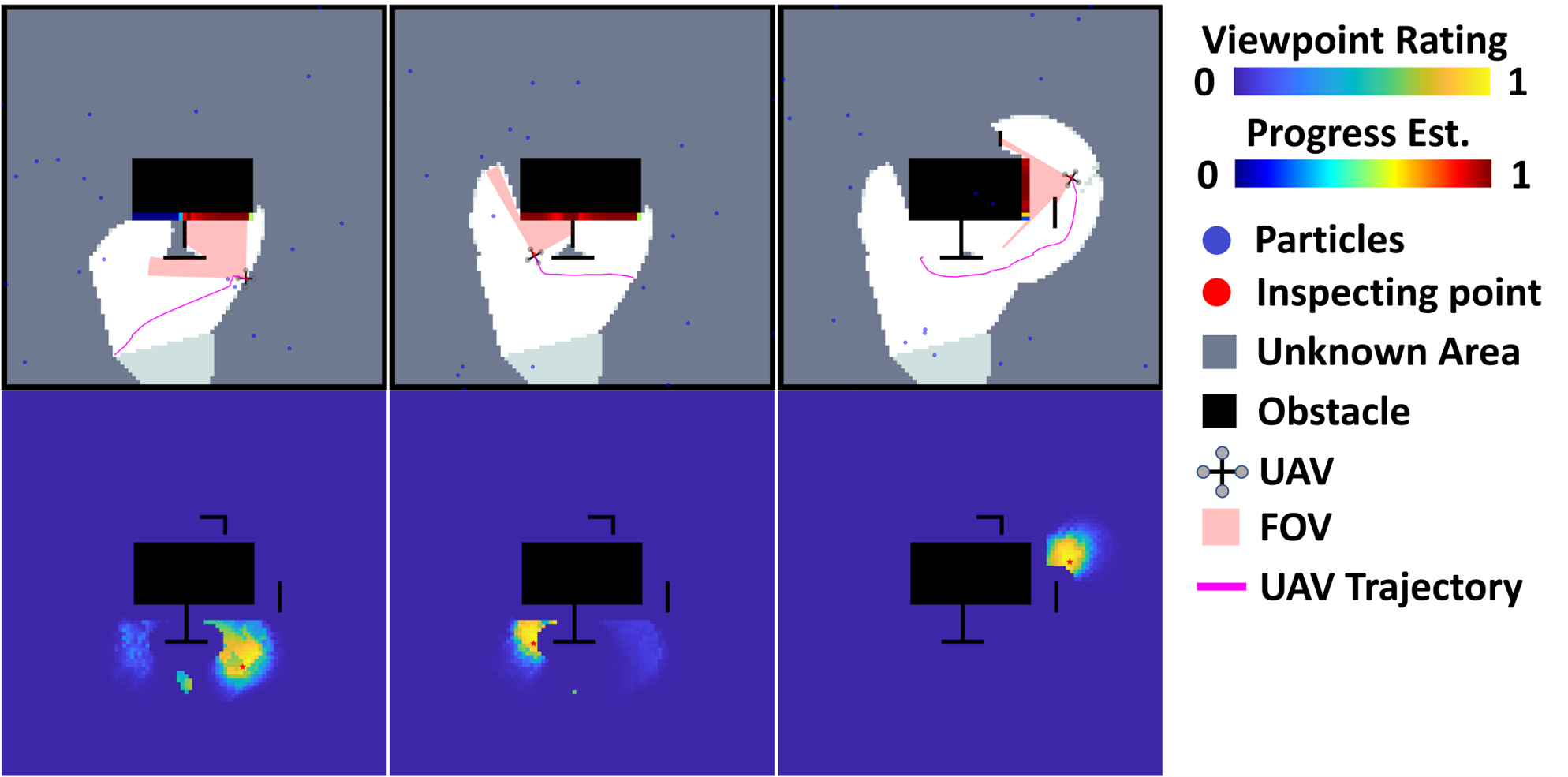}
    \vspace{-9pt}
    \caption{A 2D simulation from a top-down view, The first row displays 1) quadrotor states, 2) exploration progress, and 3) inspected target portions. The second row shows the ground truth information gain for all viewpoints.}
    \label{fig:sim2DResult}
\end{figure}

We employ Particle Swarm Optimization (PSO) with 20 particles to find the optimal viewpoint. Fig.~\ref{fig:sim2DResult} shows keyframes from the simulation and heatmaps display expected target coverage and image quality metrics from \eqref{eq:optFunc}. The L2 norm error between the globally optimal solution and the PSO results is $0.8538 \pm 0.6816$m, which is insignificant considering the size of the target.
\vspace{-10pt}
\vspace{-8pt}

\begin{figure}[h]
    \centering
    \vspace{-5pt}
    \includegraphics[width=0.95\columnwidth]{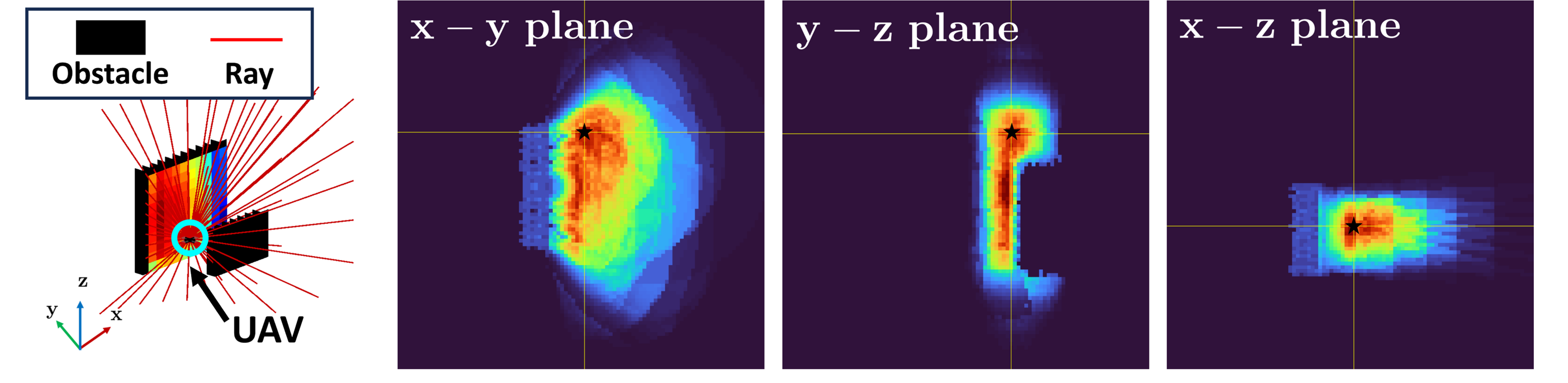}
    \vspace{-10pt}
    \caption{Heatmap showing the results of the equations in the x-y, x-z, and y-z planes, sliced at the optimal locations (denoted by a black star). As shown, the vehicle is at the optimal point and can see the whole target.}
    \label{fig:Sim3DResult}
\end{figure}

In the 3D simulation (Fig.~\ref{fig:Sim3DResult}), a target measuring 10m$\times$1m$\times$10m is positioned inside a 30m$\times$30m$\times$40m area that contains an obstacle in front. We adopt RRT$^*$ for path planning due to its effectiveness in complex 3D spaces. The aerial vehicle navigates to the optimal viewpoint while avoiding obstacles. The 3D example evaluation is conducted at a rate of 5.48ms per candidate viewpoint.
\vspace{-5pt}

\begin{figure}[h]
    \centering
    \vspace{-5pt}
    \subfigure[]{\includegraphics[width=0.287\columnwidth]{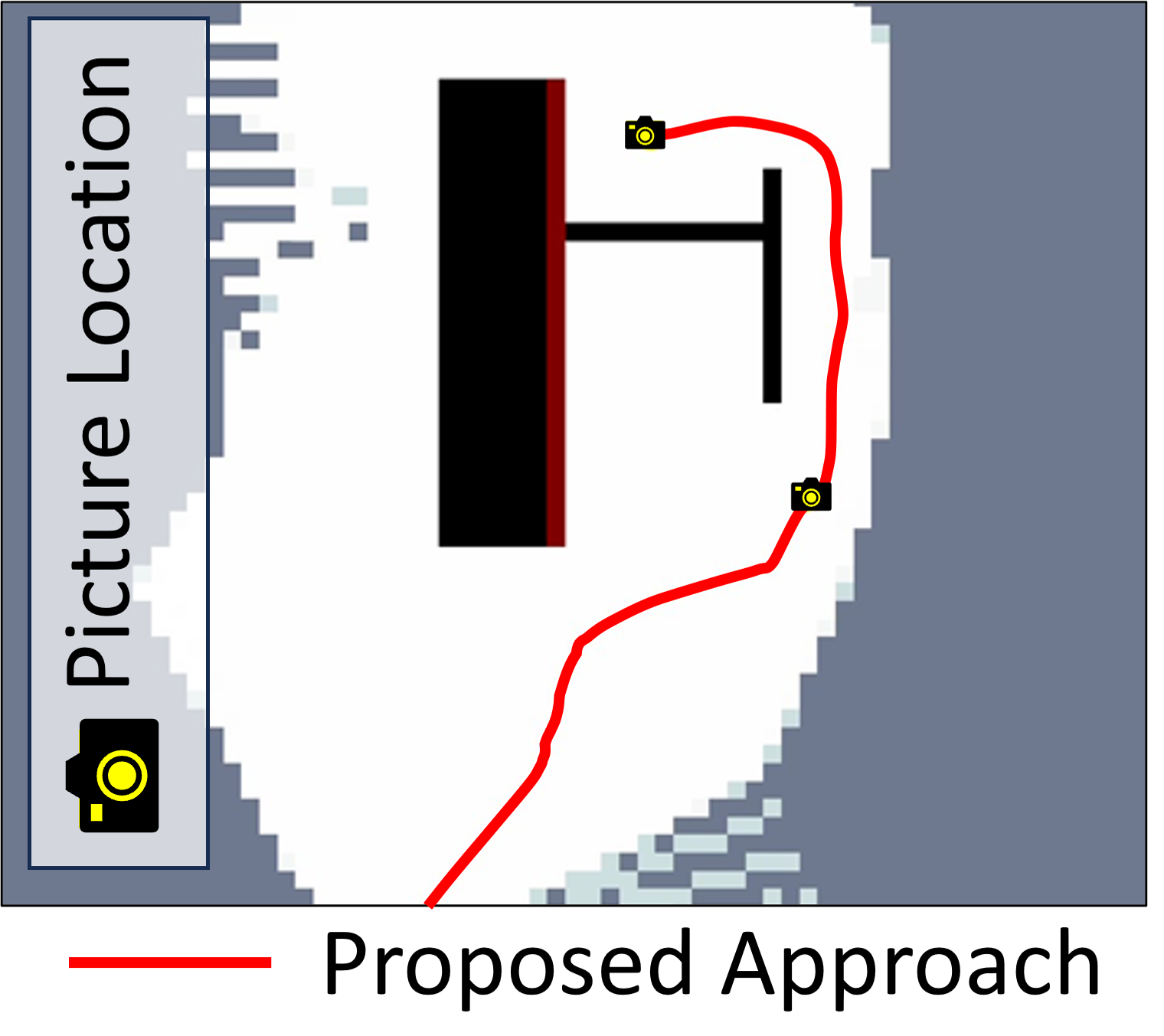}\label{fig:comparison1a}}
    \subfigure[]{\includegraphics[width=0.291\columnwidth]{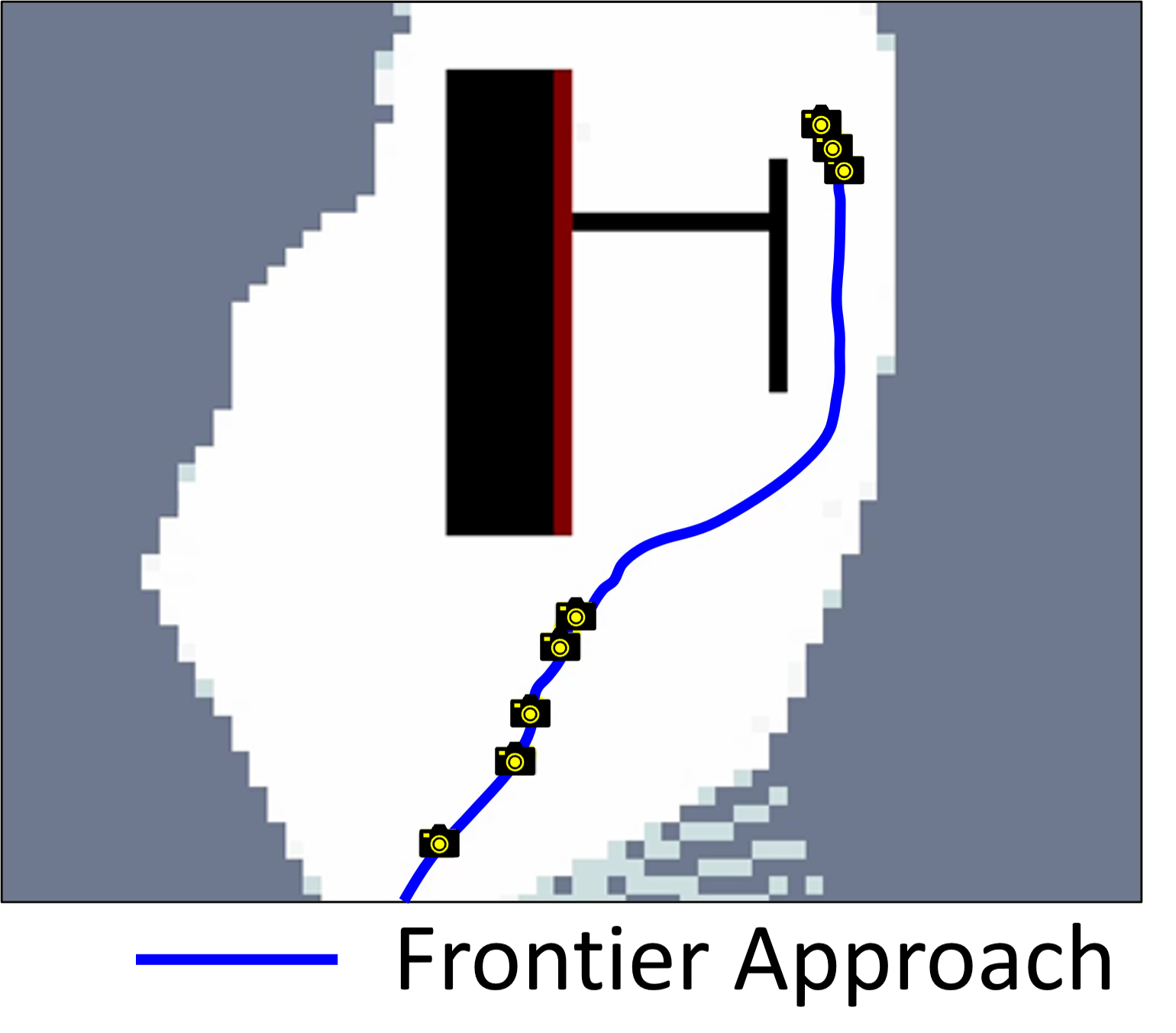}\label{fig:comparison1b}}
    \subfigure[]{\includegraphics[width=0.372\columnwidth]{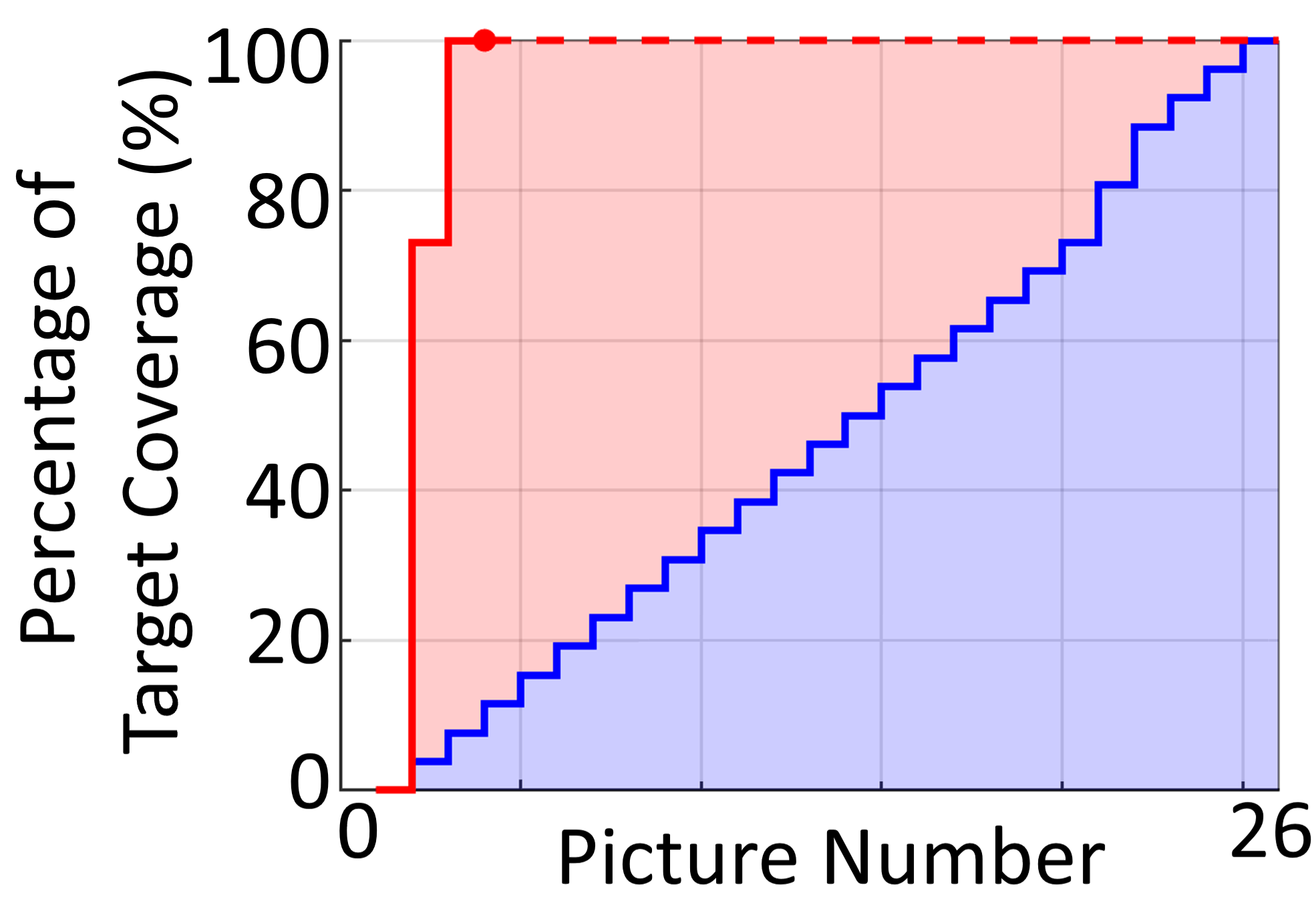}\label{fig:comparison1c}}
    \vspace{-8pt}
    \caption{Comparison of the proposed approach (a) with the sampling-based frontier approach (b).}
    \label{fig:comparison1}
\end{figure}

We note that this approach minimizes the number of photos needed and evaluates image quality using defined metrics. Fig.~\ref{fig:comparison1} illustrates a comparison between our method and a sampling-based frontier approach~\cite{respall2021fast}. While the frontier method captures 26 photos to cover the entire target, none meet our quality criteria. In contrast, our approach captures only 2 photos that capture the whole target maximizing the evaluation metrics.

\vspace{-10pt}

\section{Experiments} \label{sec:experiments}
To validate the effectiveness and versatility of our proposed approach, extensive real and virtual experiments were conducted with various autonomous vehicles. We performed 2D experiments on different robots in similar environmental settings with and without the presence of obstacles. The experiments were carried out indoors using a VICON motion capture system to localize the robots. The robots used for these experiments are: a ROSbot skid steering UGV equipped with an RPLIDAR and a camera, and a DJI Tello UAV 
operating in the x-y plane while maintaining a fixed altitude. As in the simulation, we set $\beta$ at $0.8$. 

\begin{figure}[t!]
    \centering
    \vspace{-5pt}
    \includegraphics[width=0.9\columnwidth]{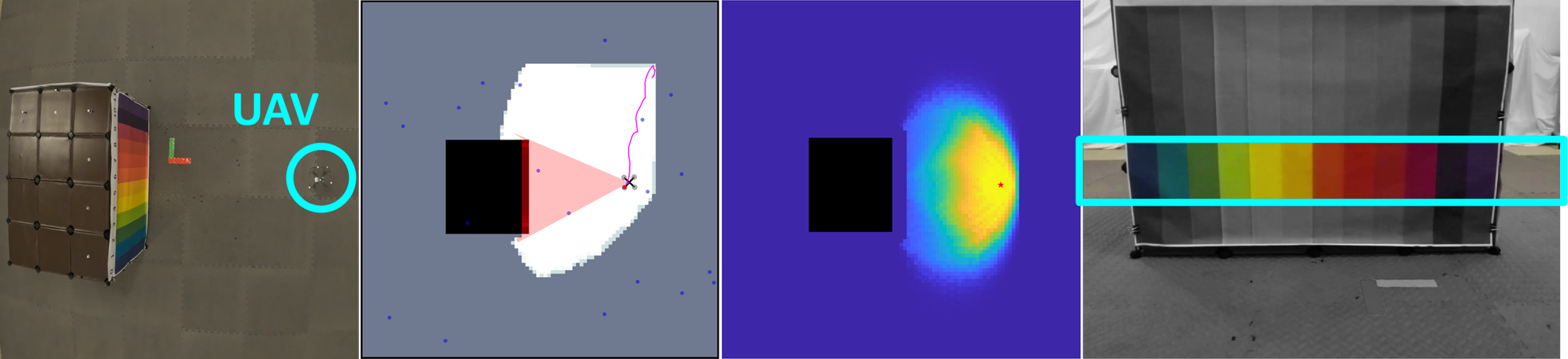}
    \vspace{-8pt}
    \caption{Without obstacles, the proposed approach enables the UAV to capture the entire target from the optimal inspecting point.}
    \label{fig:exp_no_obs}
    \vspace{-8pt}
\end{figure}

\begin{figure}[t!]
    \centering
    \includegraphics[width=0.9\columnwidth]{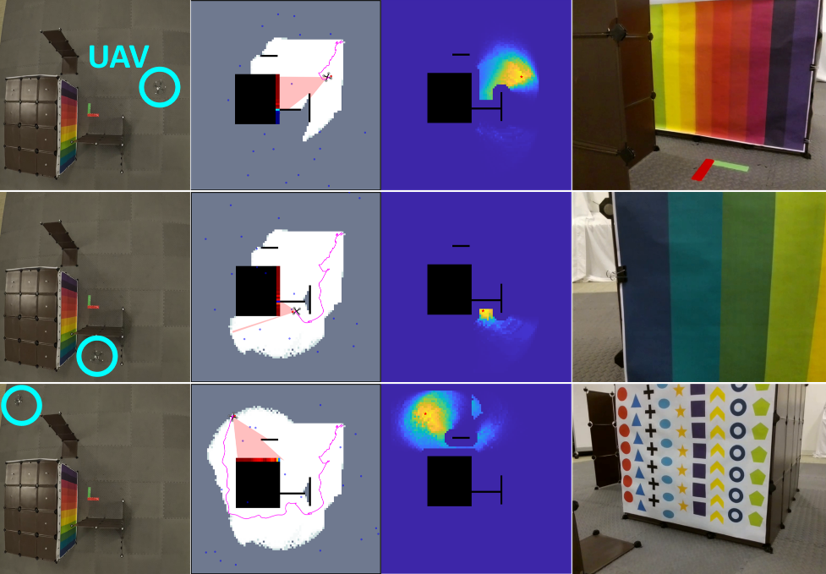}
    \vspace{-8pt}
    \caption{A 2D experiment is shown that is similar to the 2D simulation. The heatmap shows that the quadrotor captures images from the most advantageous perspectives. The photographs obtained are also displayed.}
    \vspace{-8pt}
    \label{fig:exp_with_obs}
\end{figure}

\begin{figure}[t!]
    \centering
    \includegraphics[width=0.9\columnwidth]{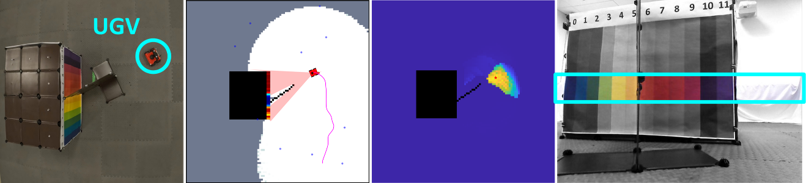}
    \vspace{-8pt}
    \caption{An experiment inspecting an object with a slash-shape obstacle. The ground vehicle preserves almost the entire target with minimal distortion. The obstacle appears as a thin line.}
    \label{fig:exp_slach_obs}
    \vspace{-5pt}
\end{figure}

First, to test the effectiveness of our approach, the Tello was started from the corner of the room and tasked to inspect an object in a free space. From the left to the right in Fig.~\ref{fig:exp_no_obs}, we show: 1) the bird view of Tello taking a picture, 2) the GP result shows the entire target is inspected, 3) the heatmap shows the ground truth evaluation of all viewpoints (note the UAV at the optimal position), and 4) the final result. Fig.~\ref{fig:exp_with_obs} shows the result for a case with obstacles blocking the direct view of the facade of a target. With our method, the UAV captures two images from each side of the T-shaped obstacle to fully cover the front surface, and only requires one angled shot to inspect the entire side surface. Lastly, Fig. \ref{fig:exp_slach_obs} shows the results for a setting with a slash-shaped obstacle. The ROSbot UGV locates the best vantage point at which the obstacle appears as a thin line.

\begin{figure}[t!]
    \centering
    \vspace{-5pt}
    \includegraphics[width=0.9\columnwidth]{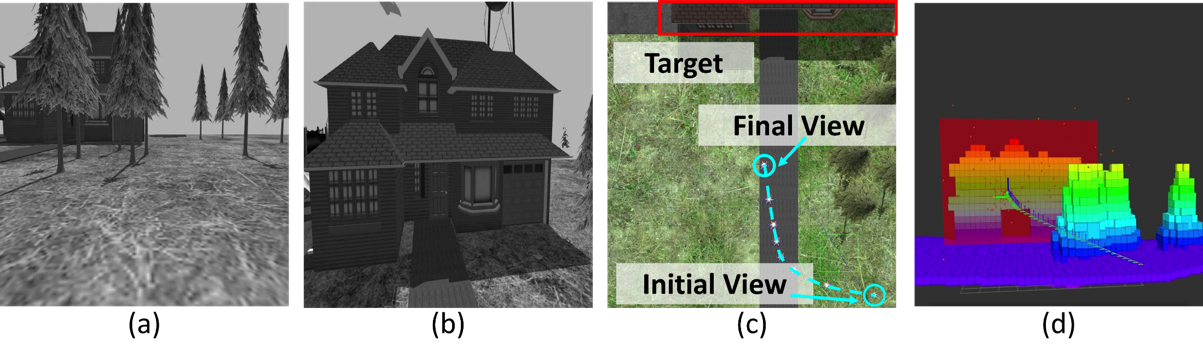}
    \vspace{-13pt}
    \caption{An example virtual experiment where a tree obstructs the view of the target. (a) and (b) show the initial and final view. (c) and (d) show the trajectory followed to the best viewpoint in the gazebo and rviz.}
    \label{fig:virtual_experiment}
    \vspace{-17pt}
\end{figure}

To further reinforce these results, we performed virtual experiments using a simulated RotorS Firefly equipped with a stereo camera. This experiment ran on the same hardware in Sec.~\ref{sec:simulation} in C++. The evaluation of each candidate viewpoint occurs in 2.39ms. As Fig.~\ref{fig:virtual_experiment} shows the results of an example virtual experiment, the vehicle detects the tree in front of the house using the stereo camera and stores the readings in an OctoMap \cite{hornung2013octomap}. The vehicle uses PSO to locate the optimal viewpoint. The picture in Fig.~\ref{fig:virtual_experiment}(b) shows the resulting image from this viewpoint. For this experiment, we set $\beta$ to 0.8 and use RRT$^*$ to route the vehicle around obstacles. The Gazebo snapshot in Fig.~\ref{fig:virtual_experiment}(c) shows the planned trajectory of the UAV. The Rviz snapshot in Fig.~\ref{fig:virtual_experiment}(d) shows the resulting plan given the observed obstacles, where the red-shaded region represents the target.

\section{Discussion and Conclusion} \label{sec:conclusion}

In this work, we have presented a novel framework for an autonomous vehicle to take a quality image of a target in a partially-known environment. The approach includes utility functions to define the quality of a viewpoint as well as a method to estimate the information gain of a viewpoint at runtime. The extensive simulations and results of the experiments show the validity, applicability, and generality of the proposed method. 

Moving forward, we are interested in the following objectives: 1) incorporate our metrics into existing NBV planners to optimally route to viewpoints; 2) extend the framework to consider dynamic obstacles; 3) identify areas of interest to inspect after the entire target has been viewed such as structural inconsistencies and areas of concern; 4) implement this approach with only a monocular camera, utilizing machine learning to recognize depth information of a target; 5) modeling and testing of this framework with multiple robots. 

\section{Acknowledgements}
This work is based on research sponsored by the CoStar Group. We would like to thank Khajamoinuddin Syed for the useful discussion about the research application. 

\bibliographystyle{IEEEtran}
\bibliography{References} 

\end{document}